\documentclass{article}
\usepackage{spconf,amsmath,graphicx}
\usepackage{subfigure}
\title{CHAM: Action Recognition Using Convolutional Hierarchical Attention Model}
\name{Shiyang Yan$^a$$^,$$^b$, Jeremy S. Smith$^a$, Wenjin Lu$^b$ and Bailing Zhang$^b$}
\address{$^a$University of Liverpool \\
$^b$Computer Science and Software Engineering Department, \\
 Xi'an Jiaotong-Liverpool University}

\begin{document}
%\ninept
\topmargin=0mm
\maketitle
\begin{abstract}
Recently, the soft attention mechanism, which was originally proposed in language processing, has been applied in computer vision tasks like image captioning. This paper presents improvements to the soft attention model by combining a convolutional  Long Short-Term Memory (LSTM) with a hierarchical system architecture to recognize action categories in videos. We call this model the Convolutional Hierarchical Attention Model (CHAM). The model applies a convolutional operation inside the LSTM cell and an attention map generation process to recognize actions. The hierarchical architecture of this model is able to explicitly reason on multi-granularities of action categories. The proposed architecture achieved improved results on three publicly available datasets: the UCF sports dataset, the Olympic sports dataset and the HMDB51 dataset.
\end{abstract}
\section{Introduction}
\label{sec:intro}

Action recognition in video has been a popular yet challenging task which has received significant attention by the computer vision society \cite{wang2013action} \cite{simonyan2014two}. The potential applications of action recognition include video retrieval (i.e., YouTube videos), intelligent surveillance and interactive systems. Compared with action recognition from still images, the temporal dynamics provides an important clue to recognize human actions in videos.

Among the proposed models to capture the spatial-temporal transition in videos, Recurrent Neural Networks (RNN) are the preferred candidate due to the special internal memory being able to process arbitrary sequences of inputs. A RNN is a class of artificial neural network where connections between the units form a directed cycle, and the internal state created from the network allows it to exhibit dynamic temporal behavior. Much research was conducted on RNNs in the 80s \cite{elman1990finding} \cite{werbos1988generalization} for time-series modeling, however this was hampered for a long period by the difficulties of training, particularly the vanishing gradient problem \cite{bengio1994learning}. Roughly speaking, the error gradients would vanish exponentially quickly with the size of the time lag between important events, which makes training very difficult. To mitigate this problem, a class of models with a long-range learning capability, called Long Short-Term Memory (LSTM), was introduced by Hochreiter, et al \cite{hochreiter1997long}. LSTM consists of memory blocks, with each block containing self-connected memory units to learn when to forget previous hidden states and when to update hidden states given new information. It has been verified that complex temporal sequences can be learnt by LSTM \cite{donahue2015long}.

LSTM has a close relationship with attention models in vision research and natural language processing (NLP). Human perception is characterized by an important mechanism of focusing attention selectively on different parts of a scene which has long been an important subject in the vision community. An attention model can be built using LSTM on top of image features to decide when the model should focus on certain parts of the image sequentially. In NLP, the attention model was proposed for sequence to sequence training in machine translation \cite{bahdanau2014neural}, where two types of attention model have been studied, hard attention and soft attention. Soft attention is deterministic and can be trained using back-propagation \cite{xu2015show}. Soft attention was then extended to the image captioning task \cite{xu2015show} since image captioning can be essentially considered as image to language translation. Sharma, et al.\cite{sharma2015action} used pooled convolutional descriptors with soft attention based models for action recognition and achieved good results. Continuing the previous research, we investigated the soft attention model in the action recognition context, and propose several improvements. Normally the LSTM is built on fully connected layers in which all the state-to-state transitions are matrix multiplication. This structure does not take spatial information into account. Xingjian, et al.\cite{xingjian2015convolutional} proposed convolutional LSTM in which all the transitions are convolutional operations. Following \cite{xingjian2015convolutional}, we improved the soft attention model based on convolutional LSTM.

In real world applications, an action is usually composed of a set of sub-actions. For instance, jump shooting basketball often consists of three sub-actions- jumping, shooting and landing. This is a typical hierarchical structure in terms of motion dynamics. In other words, actions are composed of multiple granularities. A straightforward way to model the layered action would be a hierarchical structure.
Following \cite{wang2016hierarchical} in which a Hierarchical Attention Networks (HAN) was proposed, we applied HAN with a convolutional LSTM to recognize multiple granularities of layered action categories. The proposed model can be termed CHAM which means Convolutional Hierarchical Attention Model.

Our main contributions can be summarized as follows:

(1) As deep features from CNNs preserve the spatial information, we improved the soft attention model by introducing convolutional operations inside the LSTM cell and attention map generation process to capture the spatial layout.

(2) To explicitly capture layered motion dependencies of video streams, we built a hierarchical two layer LSTM model for action recognition.

(3) We tested our model on three widely applied datasets, the UCF sports dataset \cite{rodriguez2010spatio}, the Olympic dataset \cite{niebles2010modeling} and the HMDB51 dataset \cite{Kuehne11} with improved results on other published work.

\section{Soft attention Model for Video Action Recognition}

\subsection{Convolutional Soft Attention Model}

LSTM was proposed by Hochreiter, et al \cite{hochreiter1997long} in 1997 and have subsequently been refined. LSTM is able to avoid the gradient vanishing problem and implements long term memory by incorporating memory units that allow the network to learn when to forget previous hidden states and when to update hidden states. The input, forget and output gates are composed of a sigmoid activation layer and matrix multiplication to define how much information flow should be passed to the next time-step. All the parameters in the gates can be learnt in the training process.

\begin{figure}[h]
  \centering
  \includegraphics[width=85mm]{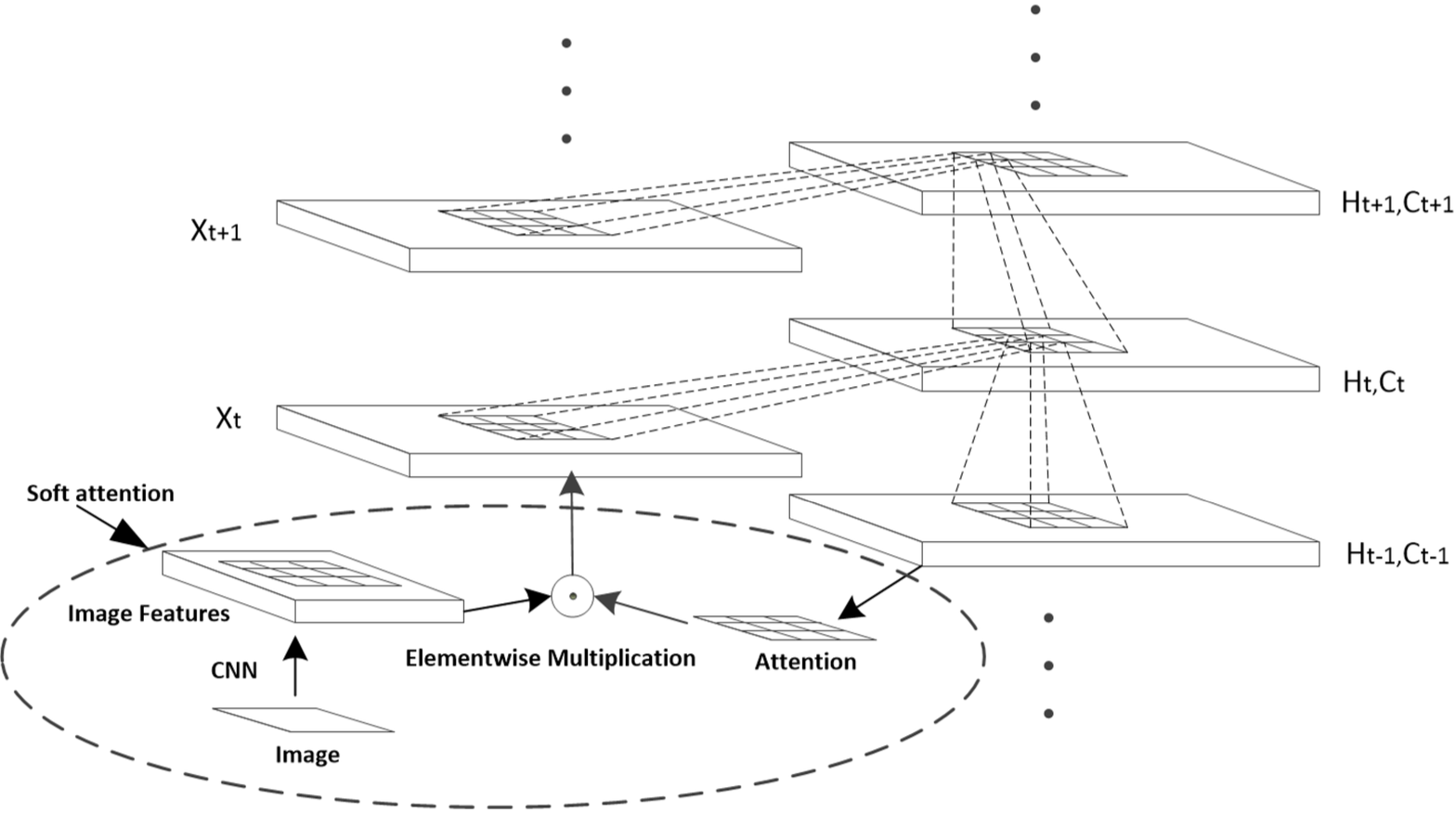}\\
  \caption{The input-to-state, and state-to-state transitions are all convolutional, the attention map is also generated by convolution. The soft attention mechanism is to elementwise multiply the attention map with image features and forward to the convolutional LSTM at each time step.}\label{softattention}
\end{figure}

Following the idea of \cite{xingjian2015convolutional}, we replaced the state-to-state transitions in LSTM with convolutional operations which are illustrated in Fig.\ref{softattention}. In Fig.\ref{softattention}, the dashed lines indicate the convolution operations, all the input-to-state and state-to-state transitions are replaced with convolutions. Moreover, the attention map is derived from the hidden layer of the LSTM also using convolutional operations. The attention map will be elementwise multiplied with image features to select the most informative regions to focus on.

Our soft attention model is built upon deep CNN features. The features were extracted from the last convolutional layer from a CNN model trained on the ImageNet \cite{deng2009imagenet} database. The last convolutional features would have shape of $K{\times}K{\times}D$. We consider the features as $K^2$ number of $D$ feature vectors in which each of the feature vectors represent overlapping receptive fields in the input image and our soft attention model choose to focus on different regions in each time step.

Letting $\sigma(x)=(1+e^{-x})^{-1}$ be the sigmoid non-linear activation function and $\phi(x)=\frac{e^x-e^{-x}}{e^x+e^{-x}}=2\sigma(2x)-1$ be the tangent non-linear activation function, the convolutional LSTM model with soft attention follows these updating rules:
\begin{equation}\label{LSTM1}
  i_t = \sigma({W_{xi}}\ast{x_t}+{W_{hi}}\ast{h_{t-1}}+b_i)
  \end{equation}
  \begin{equation}\label{LSTM2}
  f_t = \sigma({W_{xf}}\ast{x_t}+{W_{hf}}\ast{h_{t-1}}+b_f)
  \end{equation}
  \begin{equation}\label{LSTM3}
  o_t = \sigma({W_{xo}}\ast{x_t}+{W_{ho}}\ast{h_{t-1}}+b_o)
  \end{equation}
  \begin{equation}\label{LSTM4}
  g_t = \sigma({W_{xc}}\ast{x_t}+{W_{hc}}\ast{h_{t-1}}+b_c)
\end{equation}
 \begin{equation}\label{LSTM5}
  c_t = f_t \cdot c_{t-1} + i_t \cdot g_t
\end{equation}
 \begin{equation}\label{LSTM6}
  h_t = o_t \cdot \phi(c_t)
\end{equation}
Here, $i_t, f_t, o_t$ are the input, forget and output gates of the LSTM model, respectively. They are calculated according to Equations\ref{LSTM1} - \ref{LSTM3}. $c_t$ is the cell memory while $h_t$ is the hidden state of the LSTM model. A $\ast$ indicated the convolution operation. $W_\sim, b_\sim$ are convolutional weights and bias, respectively. The multiplication operations are all elementwise multiplication. $x_t$ is the input to the LSTM model at each time step. It can capture the attention information given image features and the hidden state of LSTM from the last time step. Assuming $F_t$ is the frame level image features which are $K{\times}K{\times}D$ dimension, $x_t$, the attention map on image features, can be computed as follows:
\begin{equation}\label{Attention1}
x_t = l_{t}^{ij} \cdot F_t
\end{equation}
\begin{equation}\label{Attention2}
 l_{t}^{ij} = SOFTMAX(W_z{\ast}{\phi}(W_{ha}{\ast}h_{t-1}+W_{xa}{\ast}x_{t}+b_a))
\end{equation}
$l_{t}^{ij}$ indicates the attention value of each region which is dependent on the hidden state of the last time step and the input image features of this time step. $i,j$ means the horizontal and vertical position of the attention map, respectively. We achieve this by simple weighting of the image features with attention values to preserve the spatial information instead of getting the expectation of image features as in \cite{xu2015show}. This is essentially a type of amplification of the `attention' location of features for the classification at hand. In practice, the hidden state of the last time step and input features are convolved by maps $W_{ha}$ and $W_{xa}$ respectively before passing to a softmax activation layer as in Equation \ref{Attention2}. The softmax values can be considered as the importance of each region in the image features for the model to pay attention.

Finally, the model applied the cross-entropy loss for action classification.
\begin{equation}\label{Cost}
LOSS = -\sum\limits_{t=1}^T\sum\limits_{i=1}^Cy_{t,i}log(\hat{y}_{t,i})
\end{equation}
where $y_t$ is the label vector, $\hat{y}_{t}$ is the classification probabilities at time step t. $T$ is the number of time steps and $C$ is the number of action categories.

\subsection{Hierarchical Architecture}

\begin{figure}[!t]
  \centering
  \includegraphics[width=80mm]{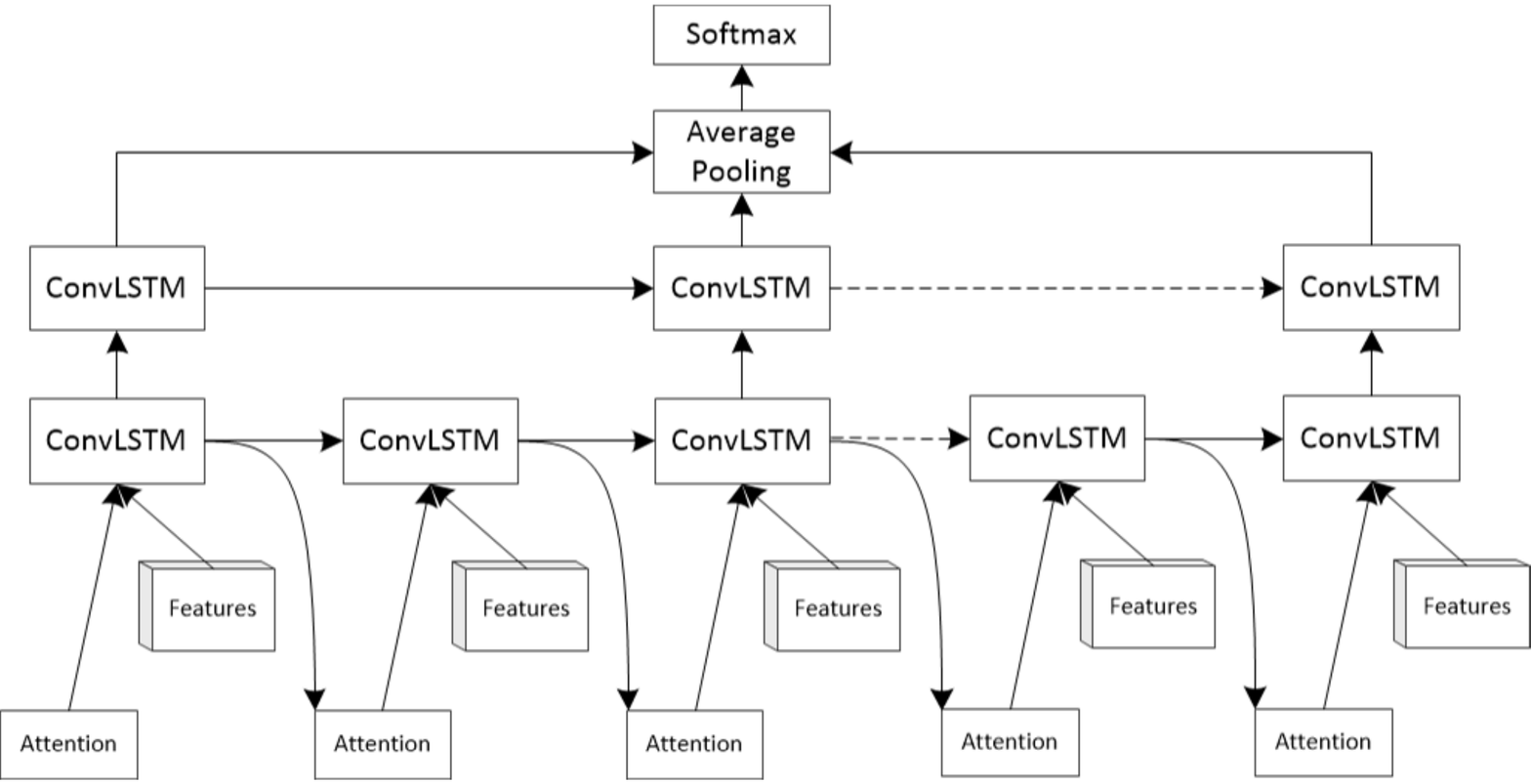}\\
  \caption{This is a two layered hierarchical model in which the first attention layer reasons on each frame and the second layer skips several steps. The outputs from the two layers are concatenated and forwarded to the average pooling layer before the softmax classifier.}\label{hiarchical}
\end{figure}

As previously introduced, the hierarchical architecture of our CHAM is to capture layered motion dependencies. Fig. \ref{hiarchical} illustrates the system structure of our hierarchical model. The first layer is the attention layer and is also able to reason on the more fine-grained properties of the temporal dependency. The second layer directly connects with first layer but skip several steps in order to catch the coarse granularity of the motion information. Then the output features of the first layer and second layer are concatenated before forwarding to the fully connected layers and an average pooling layer. Then a softmax classifier is connected to generate the results.

\section{Experiments}
\subsection{Datasets Introduction}
The approach was evaluated on three datasets, namely the UCF sports \cite{rodriguez2010spatio}, the Olympic sports \cite{niebles2010modeling} and the more difficult HMDB51 \cite{Kuehne11}. The UCF sports dataset contains actions collected from various sports on broadcast channels such as ESPN and the BBC. This dataset consists of 150 videos and with 10 different action categories present. The Olympic sports dataset was collected from YouTube sequences \cite{niebles2010modeling} and contains 16 different sports categories with 50 sequences per class. The full name of HMDB51 is Human Motion Database and it provides three train-test splits each consisting of 5100 videos. These clips are labeled with 51 action categories. The training set for each split has 3570 videos and the test set has 1530 videos.

For the UCF sports dataset, we manually divide the dataset into a training and a testing set. We used 75\% for training, and 25\% for testing. We then report the frame-level accuracy based on the testing dataset.

For the Olympic sports dataset, we used the original training-testing split with 649 clips for training and 134 clips for testing. Following \cite{niebles2010modeling}, we evaluated the Average Precision (AP) of each category on this dataset.

When evaluating our methods on HMDB51, we follow the original training-testing split and test the accuracy of each split. As \cite{sharma2015action} has the results of the conventional soft attention scheme, we only test the performance of our methodologies.

\subsection{Implementation Details}

Firstly, we extracted frame-level CNN features using MatConvNet \cite{vedaldi2015matconvnet} based on Residual-152 Networks\cite{he2015deep} trained on the ImageNet \cite{deng2009imagenet} dataset. The images were resized to 224$\times$224, hence the dimension of each frame-level features is 7$\times$7$\times$2048.

Then CHAM was built using the Theano \cite{bergstra2011theano} platform. We use a convolutional kernel size of 3$\times$3 for state-to-state transition in LSTM and a 1$\times$1 convolutional kernel for attention map generation to capture spatial information of the CNN features. When the kernel size is 3$\times$3, to ensure the states of LSTM in different time step have the same number of columns and rows as inputs, padding is needed before the convolution operation starts. All these convolutional kernels have 512 channels. A dropout is also applied on the output before being fed to the final softmax classifier with a ratio of 0.5.

Also, to carry out comparative studies, a convolutional attention model (Conv-Attention) using only one layer of the convolutional LSTM was built. The fully connected attention model (FC-Attention) based soft attention \cite{sharma2015action} was also implemented as a baseline approach. We set the matrix dimension of state-to-state transition in the fully connected LSTM as 512. The soft attention mechanism followed the settings in \cite{sharma2015action}. All the experiments were conducted using an NVIDIA TITAN X.

For the network training, we applied a mini-batch size of 64 samples at each iteration. For each video clip, the FC-Attention and Conv-Attention networks randomly selected 30 frames for training while CHAM seleted 60 frames for training with a second LSTM layer skip every 2 time steps. We applied the back propagation algorithm through time and an Adam optimizer \cite{kingma2014adam} with a learning rate of 0.0001 to train the networks. The learning rate was changed to 0.00001 after 10,000 iterations.

\subsection{Results and Discussion}

The results on the UCF sports dataset can be seen in Table \ref{UCFsports}. The Conv-Attention which apply convolutional LSTM for soft attention achieves 72\% accuracy on the UCF sports dataset while FC-attention has 70\% accuracy. CHAM has the highest accuracy of 74\% which indicates that the hierarchical architecture is able to further improve on the system performance.

\begin{table}[!t]
\caption{Accuracy on UCF sports}
 \centering
\begin{tabular}{|c|c|}
  \hline
  Methods  & Accuracy \\
  \hline
  FC-Attention   \cite{sharma2015action} & 70\% \\
  Conv-Attention(Ours)&     72\%     \\
  CHAM(Ours)   & \textbf{74\%} \\
  \hline
\end{tabular}
\label{UCFsports}
\end{table}%

\begin{table}[t]
\caption{AP on Olympics sports}
 \centering
 \resizebox{\linewidth}{!}{
\begin{tabular}{|c|c|c|c|c|c|c|}
  \hline
  Class & Vault &Triple Jump & Tennis serve&Spring board     &Snatch\\
\hline
   FC-Attention \cite{sharma2015action} & 97.0\% &88.4\%  &52.3\%  &60.0\% &23.2\%  \\

   Conv-Attention (Ours)  &97.0\%  &94.0\%    &49.8\%  &66.4\%   & 26.1 \% \\

    CHAM (Ours)   & 97.0\%   &    98.9\%    &49.5\% &69.2\% &47.8\%\\
    \hline

 Shot put &Pole vault &Platform 10m & Long jump & Javelin Throw & High jump \\
 \hline
   67.4\% &69.8\%  &84.1\%  &100.0\% &89.6\% &84.4\%  \\
    60.0\%  &100.0 \%  &86.0\% &  98.0\%   &  87.9\%  &  80.0\%      \\
    79.8\% & 60.8\% &89.7\% &    100\%  &95.0\% &78.7\%  \\
    \hline

  Hammer throw  &Discus throw &Clean and jerk &Bowling &Basketball layup & mAP\\
\hline
    38.0\% & 100.0\% &76.0\% &60.0\%  &89.8\%  &  73.7\%    \\
    36.6\%  &97.8\%  &100.0\%  &46.8\% &81.2\% &   75.5\%    \\
     37.9\%  &97.0\%  &84.8\% &46.7\%   &89.1\%    &   \textbf{76.4\%}   \\
  \hline
\end{tabular}
}
\label{Olympic}
\end{table}%

We then recorded the AP value of our methods on the Olympics sports dataset as shown in Table \ref{Olympic}. The Conv-Attention method has a mean AP value of 75.5\% which is higher than the FC-attention performance (73.7\%). Similarly, the improvement brought by the hierarchical architecture is also validated on this dataset, with a 76.4\% mean AP value achieved by the proposed CHAM model. The hierarchical model are especially good at long-term action categories, for instance, `Snatch' and `Javelin Throw' on which the CHAM method leads the other approaches by a large margin.

\begin{table}[!t]
\caption{Accuracy on HMDB51}
 \centering
\begin{tabular}{|c|c|}
  \hline
  Methods  & Accuracy \\
  \hline
  FC-Attention   \cite{sharma2015action} & 41.3\% \\
  Conv-Attention (Ours)&     42.2\%     \\
  CHAM (Ours)   & \textbf{43.4\%} \\
  \hline
\end{tabular}
\label{HMDB51}
\end{table}%

\begin{table}[!t]
\caption{Comparison with related methods on HMDB51}
 \centering
  \resizebox{\linewidth}{!}{
\begin{tabular}{|c|c|c|c|}
  \hline
  Methods  & Accuracy & Spatial Image Only & Fine-tuning \\
  \hline
  Softmax Rgression \cite{sharma2015action} & 33.5\% & Yes & No \\
  Spatial Convolutional Net \cite{simonyan2014two} & 40.5\% & Yes & Yes \\
  Trajectory-based modeling \cite{jiang2012trajectory} & 40.7\% & No & No \\
  \hline
  Average pooled LSTM \cite{sharma2015action} & 40.5\% & Yes & No \\
  FC-Attention   \cite{sharma2015action} & 41.3\% & Yes & No \\
  ConvALSTM \cite{li2016videolstm} & 43.3\% & Yes & Yes \\
  CHAM (Ours)   & \textbf{43.4\%}& Yes & No \\
  \hline
\end{tabular}
}
\label{HMDB51Cmp}
\end{table}%

The results on the HMDB51 dataset can be seen in Table \ref{HMDB51}. Similar observations can be made: the Conv-Attention has a higher accuracy value of 42.2\% and the hierarchical architecture(CHAM) added another 1.2\% gain to the final result, which is 43.4\%.

Table \ref{HMDB51Cmp} shows the comparison results on the HMDB51 dataset. From the table, the following observations can be made:

(1) Our CHAM method outperformed most of the previous methods which are only based on spatial image features.

(2) Even though our CNN model was not fine-tuned, the results still remain competitive compared with many approaches which had applied fine-tuning.

(3) The proposed model shows good potential to achieve better results. Future work can be undertaken by fine-tuning the CNN model on a specific dataset.

\begin{figure}[!t]
  \centering
  \includegraphics[width=\linewidth]{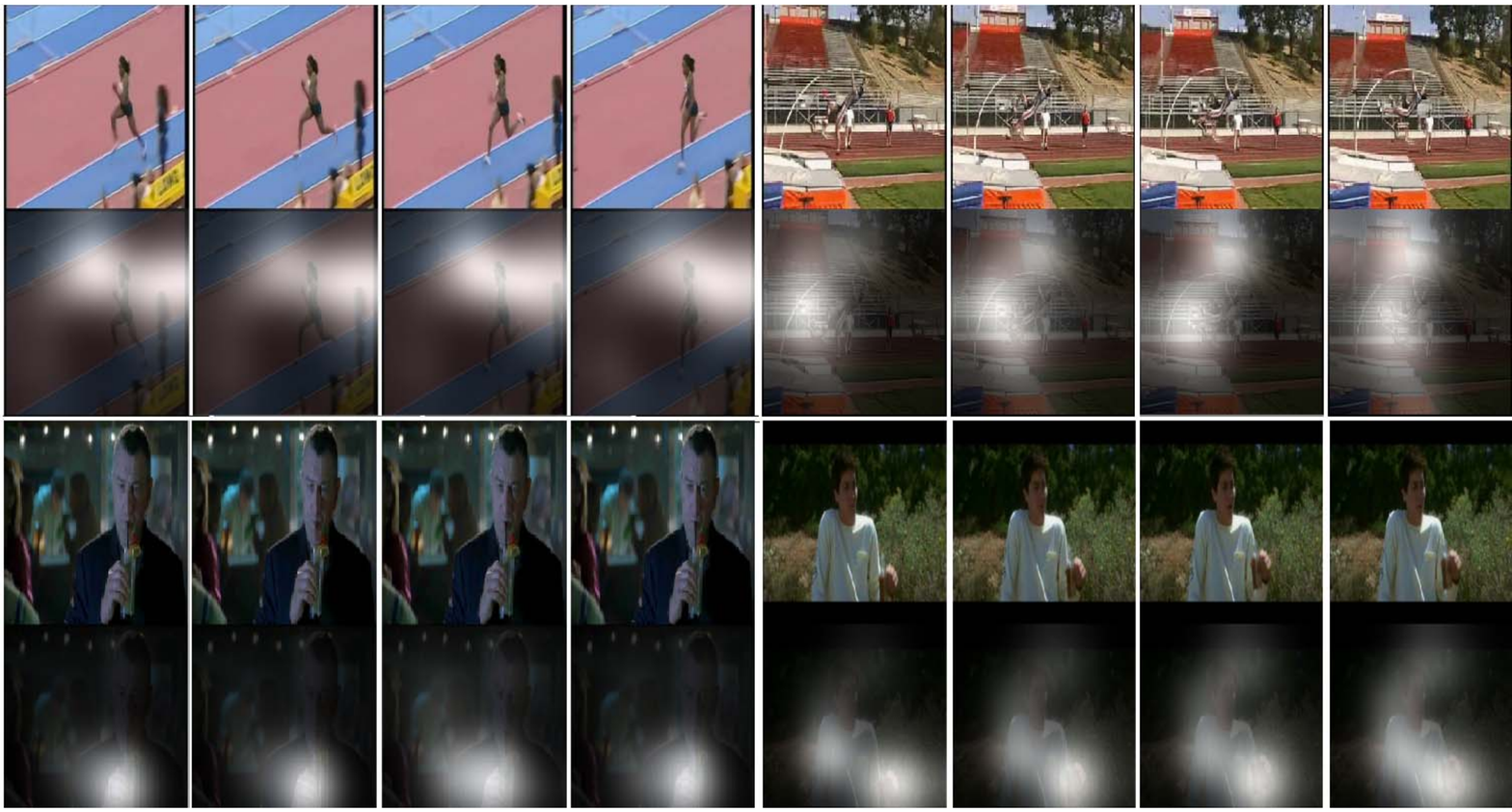}
\caption{Visualization of the attention mechanism.}\label{visualize}
\end{figure}

Fig.\ref{visualize} provide some examples of visualization of the learned attention region, we can see the regions of a person are brighter which means they are the attention region learned automatically.

\section{Conclusion}
In this paper we proposed a novel model: CHAM. This is achieved by applying convolutional LSTM, a novel RNN model, for the implementation of a soft attention mechanism and a hierarchial system architecture for action recognition. The convolutional LSTM is able to catch the spatial layout of the CNN features while the hierarchical system architecture can fuse information on the temporal dependencies from multiple granularities of the dataset. Finally, the CHAM method was tested on three widely used datasets, the UCF sports dataset, the Olympic sports dataset and the HMDB51 dataset, with improved results.

% -------------------------------------------------------------------------

\bibliographystyle{IEEEbib}
\bibliography{icip}
\end{document}